\documentclass[10pt,twocolumn,letterpaper]{article}

\usepackage{multirow}
\usepackage{multicol}
\usepackage{color}
\usepackage{subfig}
\usepackage{float}
\usepackage{iccv}
\usepackage{times}
\usepackage{epsfig}
\usepackage{graphicx}
\usepackage{amsmath}
\usepackage{amssymb}
\usepackage{setspace}
\usepackage{caption}
\usepackage[accsupp]{axessibility} 

\usepackage[breaklinks=true,bookmarks=false]{hyperref}

\iccvfinalcopy 

\def\iccvPaperID{1449} 

\ificcvfinal\fi

\begin{document}


\title{DDG-Net: Discriminability-Driven Graph Network for Weakly-supervised Temporal Action Localization}

\author{Xiaojun Tang$^{1}$, Junsong Fan$^{23}$, Chuanchen Luo$^{2}$, Zhaoxiang Zhang$^{234}$, Man Zhang$^{1}$\thanks{{\small Corresponding author.}}, and Zongyuan Yang$^{1}$\\
{\normalsize\centering$^{1}$ Beijing University of Posts and Telecommunications, China}\\
{\normalsize\centering$^{2}$ Institute of Automation, Chinese Academy of Sciences, China}\\
{\normalsize\centering$^{3}$ Centre for Artificial Intelligence and Robotics, HKISI\_CAS, HongKong, China}\\
{\normalsize\centering$^{4}$ University of Chinese Academy of Sciences, UCAS, China}\\
{\tt\small \{tangxiaojun, zhangman, yangzongyuan0\}@bupt.edu.cn,}\\
{\tt\small \{junsong.fan, luochuanchen2017, zhaoxiang.zhang\}@ia.ac.cn}
}
\maketitle
\ificcvfinal\thispagestyle{empty}\fi

\begin{abstract}
   Weakly-supervised temporal action localization (WTAL) is a practical yet challenging task. Due to large-scale datasets, most existing methods use a network pretrained in other datasets to extract features, which are not suitable enough for WTAL. To address this problem, researchers design several modules for feature enhancement, which improve the performance of the localization module, especially modeling the temporal relationship between snippets. However, all of them omit that ambiguous snippets deliver contradictory information, which would reduce the discriminability of linked snippets. Considering this phenomenon, we propose Discriminability-Driven Graph Network (DDG-Net), which explicitly models ambiguous snippets and discriminative snippets with well-designed connections, preventing the transmission of ambiguous information and enhancing the discriminability of snippet-level representations. Additionally, we propose feature consistency loss to prevent the assimilation of features and drive the graph convolution network to generate more discriminative representations. Extensive experiments on THUMOS14 and ActivityNet1.2 benchmarks demonstrate the effectiveness of DDG-Net, establishing new state-of-the-art results on both datasets. Source code is available at \url{https://github.com/XiaojunTang22/ICCV2023-DDGNet}.

\end{abstract}

\section{Introduction}

Temporal action localization (TAL) is a significant yet challenging task in video understanding. It aims to localize the start and end timestamps of the action proposals of interest and recognize their categories in untrimmed videos~\cite{MengCao2021DeepMP, KrishnaKumarSingh2017HideandSeekFA, LiminWang2017UntrimmedNetsFW}. It has attracted great attention in academia and industry due to potential applications for video retrieval, surveillance, and anomaly detection. Full-supervised temporal action localization~\cite{RuiDai2021LearningAA, ChumingLin2021LearningSB, XiaolongLiu2021EndtoendTA, DingfengShi2022ReActTA, DeepakSridhar2021ClassSA, MengmengXu2019GTADSL, ChenZhao2020VideoSG, ZixinZhu2021EnrichingLA} has made significant progress in recent years. 
\begin{figure}
\begin{center}
\includegraphics[width=1\linewidth]{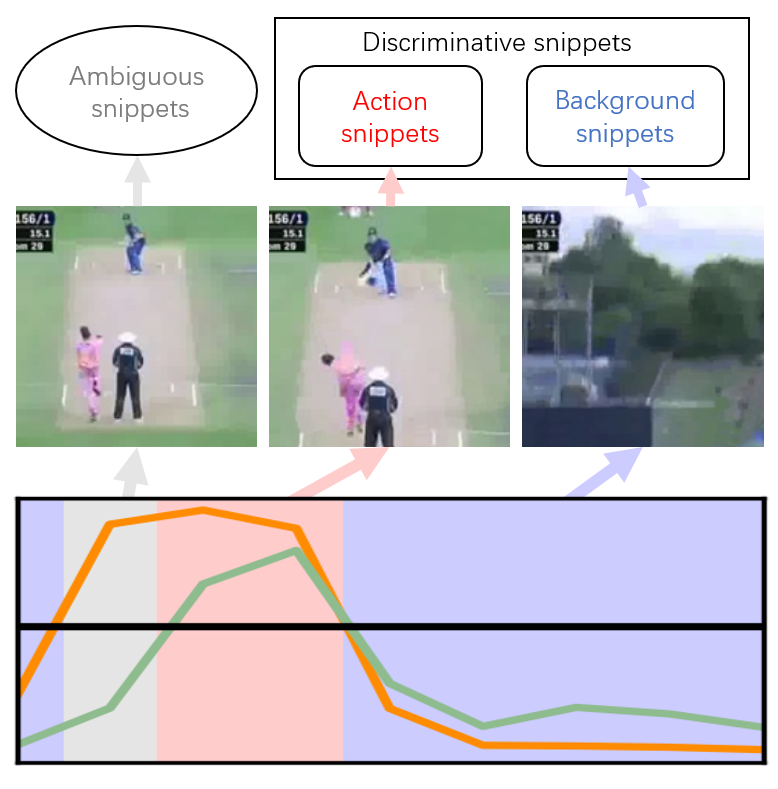}
\end{center}
   \caption{An example of ambiguous snippets. The \textcolor[RGB]{255,140,0}{orange} and \textcolor[RGB]{143,188,143}{green} curves denote the attention weights generated by RGB and optical flow separately. The black horizontal line represents the threshold.}
\label{fig:1}
\end{figure}
However, these methods require frame-level annotations, which are time-consuming and labor-intensive for large-scale datasets. Therefore, a number of researchers pay essential attention to weakly-supervised temporal action localization (WTAL).

WTAL learns to localize action instances with only video-level labels (\ie action categories in the video) during the training process. Most existing methods adopt the paradigm of multi-instance learning (MIL), regarding the video as a bag and action regions as instances. However, due to the gap between classification and localization tasks, it is difficult to localize precise action proposals without frame-level annotations. Besides, except for the discriminative snippets (action or background), there are massive ambiguous snippets (\eg action context) in untrimmed videos. For example, as shown in Figure\ref{fig:1}, during the ``CricketBowling" action, action snippets with obvious appearance and quick movement, and background snippets with irrelevant content are easy to be recognized. Instead, the ambiguous snippets (action actually) contain action information (similar appearance) and background information (slow movement) simultaneously, which increases the difficulty of discriminating.

Therefore, it is crucial to localize and recognize ambiguous snippets exactly. Several works~\cite{LinjiangHuang2020RelationalPN, ZiyiLiu2021WeaklyST, ZiyiLiu2021ACSNetAS,  SanqingQu2021ACMNetAC} are devoted to relieving the interference of action context by separating action and context.
Meanwhile, we note that a few works ~\cite{BoHeASMLocAS, MaheenRashid2020ActionGW, HaichaoShiDynamicGM, ZichenYangACGNetAC} pursue to model the temporal relationship between snippets for receiving more complete information. These methods have achieved certain performances, but they overlook the adverse effects of ambiguous information. When ambiguous snippets spread out information to similar action or background snippets, ambiguity is delivered together. 
It leads to the result that ambiguous snippets would reduce the discriminability of others when transmitting contradictory information to them. 

Inspired by the above research and analysis, we propose a novel graph network, named Discriminablity-Driven Graph Network (DDG-Net), which explicitly separates ambiguous snippets and discriminative snippets to perform more effective graph inference with well-designed connections. Specifically, as shown in Figure \ref{fig:1}, we divide firstly the whole video into three types of snippets (\ie pseudo-action, pseudo-background, and ambiguous snippets). Then, we design different connections among them to ensure only the transmission of discriminative information. In this way, the snippets receive complementary information from linked snippets through graph inference. What is more, the discriminability of ambiguous snippets is enhanced while the snippets never do harm to other snippets due to distinctive connections. Besides, we propose feature consistency loss to maintain characteristics of snippet-level representations for localization. Our main contributions can be summarized as follows.

\begin{itemize}
    \item We propose a novel graph network DDG-Net, which explicitly models ambiguous and discriminative snippets with different types of connections, aiming to spread complementary information and enhance the discriminability of snippet-level features while eliminating the adverse effects of ambiguous information. 
\end{itemize}
\begin{itemize}
    \item We propose feature consistency loss with DDG-Net, which prevents the assimilation of snippet-level features and drives the graph convolution network to generate more discriminative representations.
\end{itemize}
\begin{itemize}
    \item Extensive experiments demonstrate that our method is effective, and establishes new state-of-the-art results on THUMOS14 and ActivityNet1.2 datasets.
\end{itemize}
\section{Related Work}
{
\setlength{\parindent}{0cm}
\textbf{Weakly-supervised temporal action localization.} In recent years, Temporal action localization with weak supervision has attracted increasing interest from the community. Weak supervision includes single-frame annotations~\cite{XinpengDing2020WeaklyST,ChenJu2021DivideAC,PilhyeonLee2021LearningAC, FanMa2020SFNetSS, DavideMoltisanti2019ActionRF, LeYang2021BackgroundClickSF} and video-level labels~\cite{AshrafulIslam2020WeaklyST,ChenJu2021AdaptiveMS, SanathNarayan20193CNetCC, RunhaoZeng2019BreakingWI, ChengweiZhang2019AdversarialSS, JiaXingZhong2018StepbystepEO}, and we focus on the latter. 
UntrimmedNet~\cite{LiminWang2017UntrimmedNetsFW} firstly solves the task via a classifier module and a selection module. 
STPN~\cite{PhucXuanNguyen2018WeaklySA} exploits the sparse loss to suppress unimportant selections. 
W-TALC~\cite{SujoyPaul2018WTALCWT} punishes the distance between video-level features with the same action class. 
Bas-Net~\cite{PilhyeonLee2019BackgroundSN} introduces background class as an auxiliary to propose a two-branch architecture and suppress activation of background snippets. 
~\cite{BoHeASMLocAS, HaishengSu2018CascadedPM} generate more complete snippet-level representations by modeling temporal relationships.
~\cite{MihirJain2020ActionBytesLF, HaishengSu2021TransferableKM, XiaoyuZhang2019AdapNetAD} learn extra knowledge from trimmed videos.
DELU~\cite{MengyuanChenDualEvidentialLF} introduces Evidential Learning which effectively alleviates ambiguity between action and background. 
~\cite{ZiyiLiu2021WeaklyST, ZiyiLiu2021ACSNetAS, SanqingQu2021ACMNetAC} make efforts to separate action and background by modeling action context.
~\cite{JunyuGaoFinegrainedTC, li2022exploring, CanZhang2021CoLAWT} adopt contrastive learning for action-background seperation.
~\cite{YuanhaoZhai2020TwoStreamCN} proposes a two-stream network to eliminate false positive action proposals. 
CO2-Net~\cite{FaTingHong2021CrossmodalCN} exploits cross-modal channel attention to filter out redundant information.
~\cite{YuanJi2021WeaklySupervisedTA, WenfeiYang2021UncertaintyGC} aim to train a stronger model via collaborative learning. These methods exploit the collaboration of RGB and optical flow which is also significant for our model. 
}

As mentioned above, most existing methods have devoted significant efforts to improving localization modules. Different from them, we focus on modeling more suitable features for WTAL. Specifically, we generate more discriminative features by transferring complementary information and eliminating ambiguity via a graph network.

{
\setlength{\parindent}{0cm}
\textbf{Graph-based Weakly-supervised temporal action localization.} Rashid \etal~\cite {MaheenRashid2020ActionGW} constructs sparse action graphs by treating snippets as nodes to connect relevant snippets. 
DGCNN~\cite{HaichaoShiDynamicGM} constructs a dynamic graph with various relations in order to make full use of temporal and contextual relations between pseudo-action snippets.
ACGNet~\cite{ZichenYangACGNetAC} spreads the complementary information through graph inference.
 Though these works enhance snippet features via graph networks, they ignore the effects of ambiguous information which may reduce the discriminability of snippet-level representations.
}

Unlike these graph networks, We subdivide the whole video into discriminative and ambiguous snippets to construct different types of connections with the consideration that ambiguous snippets would confuse linked snippets. Specially, we force ambiguous snippets to never spread information to others and only receive relevant information from discriminative snippets. In this way, we enhance the discriminability of ambiguous snippets and prevent the propagation of ambiguity.

\section{Base Model}
In this section, we introduce the base model based on DELU~\cite{MengyuanChenDualEvidentialLF} for WTAL.
We first define the problem formulation (Section \ref{3.1}). Then we give the pipeline with the base model (Section \ref{3.2}).
\subsection{Problem Formulation}\label{3.1}
Given an untrimmed video \(\mathbb{V}\) and its corresponding multi-hot action category label \(y\in\{0, 1\}^{C}\) for training, where \(C\) is the number of action categories, the goal of temporal action localization is to get a set of action instances \(S={(s_{i},e_{i},c_{i},\varphi_{i})}\) in a testing video, where \(s_{i}\) and \(e_{i}\) is the start and end timestamps of the corresponding action instance, \(c_{i}\) and \(\varphi_{i}\) is the predicted action category and related confidence score.
\subsection{Pipeline}\label{3.2}
Following previous works~\cite{MengyuanChenDualEvidentialLF, FaTingHong2021CrossmodalCN, PilhyeonLee2019BackgroundSN}, we first divide an untrimmed video into \(T\) non-overlapping 16-frame snippets. Then, snippet-level features are extracted from RGB and optical flow streams by pretrained networks. For convenience, we denote snippet features as \(F^{*}=[f_{1}^{*}, ..., f_{T}^{*}]\in\mathbb{R}^{D\times T}\), where \(D\) is the dimension of feature and \(*\in\{r, f\}\) refers to RGB and optical flow.

We remove the cross-modal consensus module~\cite{FaTingHong2021CrossmodalCN} from DELU~\cite{MengyuanChenDualEvidentialLF} to emphasize our method because both are modules for feature enhancement. 
We present extra results combined with the cross-modal consensus module in the supplementary material.

After \(F^{*}\) are extracted, they are put through an attention module to generate action attention weights as follows,
\begin{equation}
\mathbb{A}^{*}=\Phi^{*}(F^{*})\in \mathbb{R}^{1\times T}
\end{equation}
\(\Phi\) is the attention module consisting of several convolution layers and a sigmoid function. We fuse two attention sequences for use later.
\begin{equation}
\mathbb{A}=\frac{\mathbb{A}^{r}+\mathbb{A}^{f}}{2}
\end{equation}
Then we generate a classification activation sequence (CAS) through a feature fusion module and a classifier, formally,
\begin{equation}
p=\Psi(\Theta(F^{r} \oplus F^{f}))\in \mathbb{R}^{(C+1)\times T}
\end{equation}
\(\oplus\) is a concatenate operator through the channel dimension. \(\Theta\) is the feature fusion module and \(\Psi\) is the classifier include \(C\) action classes and background class.  We further get the suppressed CAS \(\overline{p}=\mathbb{A} \cdot p\), where \(\cdot\) denotes element-wise multiplication.

We denote the train objective function of the base model as \(L_{base}\).
At the testing stage, we follow the standard process~\cite{MengyuanChenDualEvidentialLF} to carry out temporal action localization.

\section{Method}
\begin{figure*}[htb]
\begin{center}
\includegraphics[width=1\linewidth]{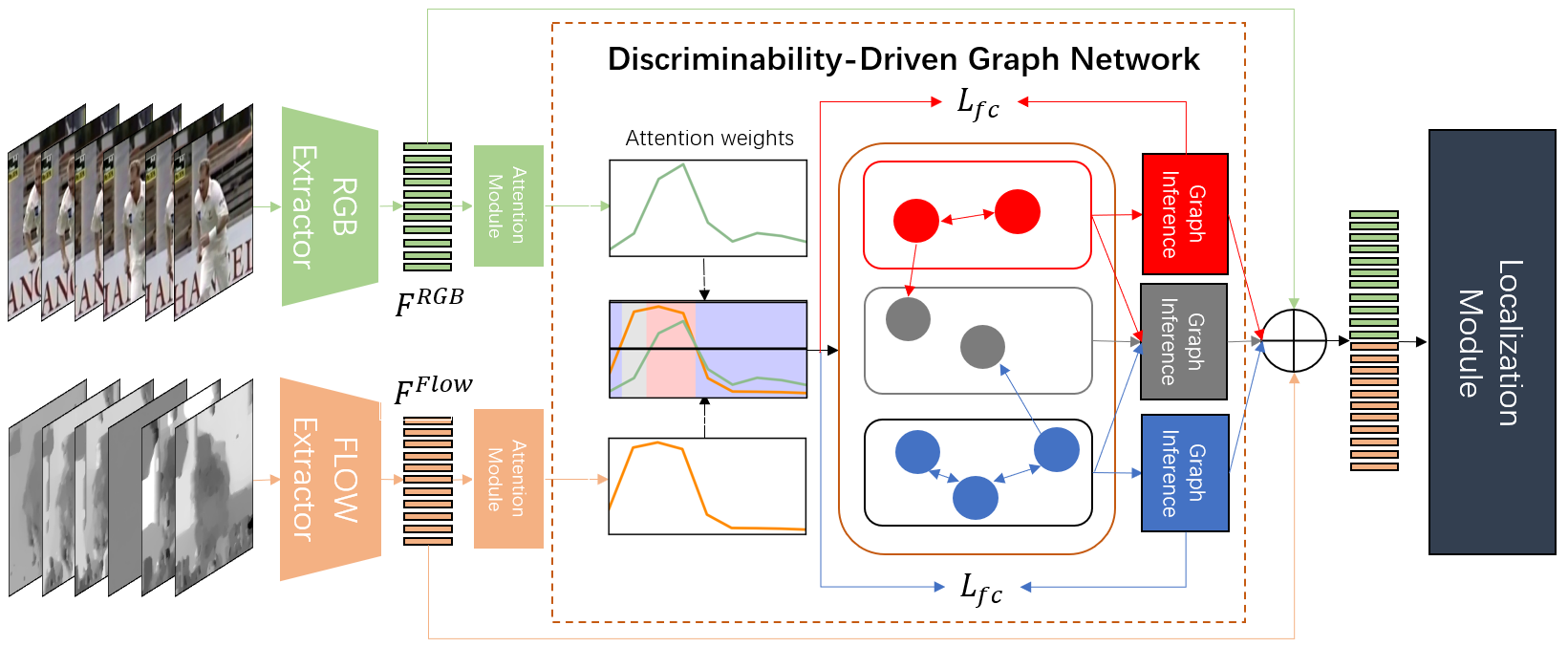}
\end{center}
   \caption{Overview of the proposed DDG-Net. Based on action weights, the snippets are pre-classified as \textcolor{red}{pseudo-action}, \textcolor{blue}{pseudo-background}, and \textcolor[RGB]{120,120,120}{ambiguous} snippets respectively. Different connections are linked among them to construct three subgraphs. More discriminative features are generated via graph inference for the localization module.}
\label{fig:2}
\end{figure*}
In this section, we describe our approach in detail. The architecture of DDG-NeT is shown in Figure \ref{fig:2}. We first give definitions of graphs (Section \ref{4.1}). Then we introduce the composition principle of graphs (Section \ref{4.2}). And the complementary and discriminative information is spread out via graph inference (Section \ref{4.3}). To prevent the assimilation of snippet-level features, we propose feature consistency loss which maintains characteristics of discriminative representations (Section \ref{4.4}). 
It is worth mentioning that our method can be applied to any two-stream model~\cite{MengyuanChenDualEvidentialLF, FaTingHong2021CrossmodalCN, YuanhaoZhai2020TwoStreamCN}.
\subsection{Graph Formulation}\label{4.1}
We denote the complete graph as \(G=(V, E)\), and \(A\) is its adjacency matrix. In this case, \(V\) denotes the set of all snippets in the whole video. We divide it into three subsets \(V_{a}\), \(V_{b}\), \(V_{m}\), which contain pseudo-action, pseudo-background and ambiguous snippets respectively. \(V_{a}\) with symmetrical adjacency matrix \(A_{a}\) constructs a undirected graph \(G_{a}\), and we regard it as action graph. Background graph \({G_{b}}\) is also constructed by \(V_{b}\) with \(A_{b}\). The nodes of ambiguous graph \(G_{m}\) consisting of \(V\) is a directed graph, and the direction of edges is always from nodes of \(V_{a}\) or \(V_{b}\) to \(V_{m}\). Under such a setting, we construct two subgraphs for interaction between relevant discriminative snippets and a subgraph for receiving discriminative information.

\subsection{Graph Generation}\label{4.2}
We first introduce the pre-classification method for snippets. Without frame-level annotations, we have no supervision to snippet-level categories. Learning from the previous method~\cite{ZiyiLiu2021WeaklyST}, we consider the action weights of snippets as the judgment basis. We jointly take \(\mathbb{A}^{r}\) and \(\mathbb{A}^{f}\) into consideration. For each timestamp \(t\), when \(\mathbb{A}^{r}_{t}\) and \(\mathbb{A}^{f}_{t}\) exceed the action threshold \(\eta\), it is regarded as a pseudo-action snippet. Also when \(\mathbb{A}^{r}_{t}\) and \(\mathbb{A}^{f}_{t}\) be lower than threshold 1-\(\eta\), it is regarded as a pseudo-background snippet. If a snippet neither belongs to both, we regard it as an ambiguous snippet.

We simply treat feature similarity as the edge weights between connected nodes. Specially, we construct the graph \(G^{*}\) and compute its adjacency matrix \(A^{*}\) separately for RGB and optical flow features. Exploiting the collaboration of RGB and optical flow, we take the average of their adjacency matrixes instead of the respective adjacency matrix for graph inference. Formally,
\begin{equation}
A_{ij}^{*}=s(f^{*}_{i},f^{*}_{j})
\end{equation}
\begin{equation}\label{fuse}
A_{ij} = \frac{A_{ij}^{r} + A_{ij}^{f}}{2}
\end{equation}
\(s\) is the similarity measure, cosine similarity in our experiments.

In order to write concisely, we appropriately omit the logo of RGB and optical flow in the following. We construct three subgraphs according to the division of snippets.

The values of adjacency matrix \(A_{a}\) and \(A_{b}\) are consistent with the values on the corresponding position of \(A\). The ambiguous graph \(G_{m}\) is a directed graph, whose adjacency matrix \({A}_{m}\) is masked partly. Formally,
\begin{equation}
\begin{split}
A_{m}(i,j)= \left \{
\begin{array}{ll}
    A(i,j),                    & j\in V_{m} \quad  and \quad  i \notin V_{m};\\
    1,     & i=j \quad  and \quad  i \in V_{m};\\
    0,                                 & otherwise.
\end{array}
\right.
\end{split}
\end{equation}
Under this treatment, we exploit complementary information of discriminative snippets and abandon confusing information from ambiguous snippets.
With different types of connections, the discriminative information is delivered to relevant snippets while the ambiguous is never spread out. Note that the fusion of cross-modal adjacency matrixes and preventing the propagation of ambiguous information is vital for our model (Section \ref{5.3}).

Following ~\cite{ZichenYangACGNetAC}, we use a similarity threshold \(\theta\) and a top-K ranking list to filter out nodes with a low contribution. A softmax operator through columns is used to obtain the final adjacency matrix.

\subsection{Graph Inference}\label{4.3}
For action graph and background graph, graph average and graph convolution are directly operated to obtain aggregated features following residual connection. Taking action graph as an example, graph average is applied to action features \(F_{a}\) as follows,
\begin{equation}
F_{a}^{avg}=F_{a}A_{a}
\end{equation}
For a graph convolution network (GCN) with \(L\) layers, the formula on the \(l\)-th layer is presented,
\begin{equation}
F_{l}^{gcn}=\sigma(W_{a}^{l}F_{l-1}^{gcn}A_{a})
\end{equation}
where \(F_{l}^{gcn}\) is output of the \(l\)-th layer, \(F_{0}^{gcn}=F_{a}\), \(\sigma\) is the activation function LeakyReLU with parameter of 0.2. The aggregated features \(F_{a}^{g}\) through the graph network is the mean of \(F_{a}^{avg}\) and \(F_{a}^{gcn}\)(\ie \(F_{L}^{gcn}\)) and enhanced features are got as follows, 
\begin{equation}\label{e9}
F_{a}'=\frac{F_{a}^{g}+F_{a}}{2}
\end{equation}

For ambiguous graph, the graph average is applied as above. We never apply a separate GCN for ambiguous features due to the difficulty of training. Instead, we exploit \(F_{a}^{gcn}\) and \(F_{b}^{gcn}\) to aggregate representations of same forms. We denote \(A_{m, a}\) as the partition of \(A_{m}\) which only contain ambiguous columns and action rows. \(A_{m, b}\), \(A_{m, m}\) is the same. Formally,
\begin{equation}
F_{m}^{gcn}=F_{a}^{gcn}A_{m,a}+F_{b}^{gcn}A_{m,b}+F_{m}A_{m,m}
\end{equation}
Note that \(A_{m,m}\) is a diagonal matrix. The subsequent process is consistent with the equation \ref{e9}. In this case, the GCN only learns to transform discriminative snippet features. Ambiguous snippet representations benefit from the enhanced features via graph inference.

Through graph inference, we get enhanced snippet-level features. We sort updated features together by time and input them instead of original features into the base model to carry out the WTAL task. We find that directly using the GCN is harmful because it drives the model to pursue classification performance, resulting in poor localization with chaotic features. So we propose feature consistency loss to address the problem effectively. 
\subsection{Feature Consistency Loss}\label{4.4}
In our experiment, the performance of DDG-Net suffers a sharp decline without any constraint to the GCN. The GCN prefers to transform all snippet-level features close to action characteristics for the classification task, which is harmful to the localization task. Therefore, we propose a feature consistency loss to avoid this case. Intuitively, we hope the most discriminative representations remain unchanged through DDG-Net because they can be recognized easily. However, it is difficult to achieve this goal under the influence of graph interaction. Instead, We punish the distance between graph average features and graph convolution features to realize the same function. Formally,
\begin{equation}
\begin{split}
L_{fc} = \frac{1}{|V_{a}|}\sum\limits_{t\in V_{a}} w(\mathbb{A}_{t})d(F_{a,t}^{gcn},F_{a,t}^{avg})\\
+\frac{1}{|V_{b}|}\sum\limits_{t\in V_{b}} w(1-\mathbb{A}_{t})d(F_{b,t}^{gcn},F_{b,t}^{avg})
\end{split}
\end{equation}
\begin{equation}
w(x)=exp(-(\frac{1}{x}-1)/\tau)
\end{equation}
where \(w(\cdot)\) is a weight function and \(\tau\) is the temperature parameter. \(d(\cdot)\) is the distance measure, euclidean distance in our experiments. Under this constraint, the GCN learns to enhance less discriminative features while maintaining characteristics of the most discriminative representations. 

\subsection{Train Objective}
{
\setlength{\parindent}{0cm}
The complete train objective function is 
\begin{equation}
L=L_{base}+\lambda_{1}L_{fc}+\lambda_{2}L_{cl}
\end{equation}
where \(L_{cl}\) is the complementary learning loss.~\cite{JiaRunDu2022WeaklySupervisedTA}
}

\begin{table*}[htb]
\begin{center}
\resizebox{\linewidth}{!}{
\begin{tabular}{|c|l|cccccccc|}
\hline
\multirow{2}{*}{Supervision} & \multirow{2}{*}{Method} & \multicolumn{8}{c|}{mAP(\%)@IoU} \\
                             &                          & 0.1 & 0.2 & 0.3 & 0.4 & 0.5 & 0.6 & 0.7 & Avg\\ \hline
\multirow{4}{*}{\begin{tabular}[c]{@{}c@{}}Fully\end{tabular}}
        &S-CNN~\cite{shou2016temporal}&47.7&43.5&36.3&28.7&19.0&10.3&5.3&27.3\\
        &TAL-Net~\cite{chao2018rethinking}&59.8&57.1&53.2&48.5&42.8&33.8&20.8&45.1\\
        &DBS~\cite{gao2019video}&56.7&54.7&50.6&43.1&34.3&24.4&14.7&39.8\\
        &AGCN~\cite{li2020graph}&59.3&59.6&57.1&51.6&38.6&28.9&17.0&44.6\\
        \hline
\multirow{3}{*}{\begin{tabular}[c]{@{}c@{}}Weakly\\ single-frame\end{tabular}} 
&SF-Net~\cite{FanMa2020SFNetSS} &68.3 &62.3& 52.8& 42.2& 30.5& 20.6& 12.0 &41.2\\
&Ju \etal~\cite{ChenJu2021DivideAC}&72.8& 64.9& 58.1& 46.4 &34.5 &21.8 &11.9 &44.3\\
&Lee and Byun~\cite{PilhyeonLee2021LearningAC}&75.7& 71.4& 64.6 &56.5& 45.3& 34.5 &21.8&52.8\\
\hline
\multirow{16}{*}{\begin{tabular}[c]{@{}c@{}}Weakly\\ video-level\end{tabular}}

        &STPN~\cite{PhucXuanNguyen2018WeaklySA} & 52.0& 44.7& 35.5& 25.8& 16.9& 9.9& 4.3& 27 \\ 
        &BaS-Net~\cite{PilhyeonLee2019BackgroundSN} & 58.2& 52.3& 44.6& 36.0& 27.0& 18.6& 10.4& 35.3\\ 
        &ACM-Net~\cite{SanqingQu2021ACMNetAC} & 68.9& 62.7& 55.0& 44.6& 34.6& 21.8& 10.8& 42.6 \\ 
        &CoLA~\cite{CanZhang2021CoLAWT}& 66.2& 59.5& 51.5& 41.9& 32.2& 22.0& 13.1& 40.9 \\ 
        &CO2-Net~\cite{FaTingHong2021CrossmodalCN} & 70.1& 63.6& 54.5& 45.7& 38.3& 26.4& 13.4& 44.6 \\
        &CSCL~\cite{YuanJi2021WeaklySupervisedTA} &68.0&61.8&52.7&43.3&33.4&21.8&12.3&41.9\\ 
        &FAC-Net~\cite{LinjiangHuang2021ForegroundActionCN} &67.6 &62.1 &52.6 &44.3 &33.4 &22.5 &12.7 &42.2\\
        &ACGNet~\cite{ZichenYangACGNetAC}&68.1&62.6&53.1&44.6&34.7&22.6&12.0&42.5\\ 
        &ASM-Loc~\cite{BoHeASMLocAS} & 71.2 &65.5& 57.1& 46.8 &36.6& 25.2 &13.4& 45.1 \\
        &FTCL~\cite{JunyuGaoFinegrainedTC} &69.6&63.4&55.2&45.2&35.6&23.7&12.2&43.6\\
        &DCC~\cite{li2022exploring} &69.0&63.8&55.9&45.9&35.7&24.3&13.7&44.0\\
        &Huang \etal\cite{LinjiangHuangWeaklyST} &71.3&65.3&55.8&47.5&38.2&25.4&12.5&45.1\\
        &DGCNN~\cite{HaichaoShiDynamicGM}&66.3&59.9&52.3&43.2&32.8&22.1&13.1&41.3\\

        &DELU~\cite{MengyuanChenDualEvidentialLF} & 71.5 & 66.2 & 56.5 & 47.7 & 40.5 &27.2 & {\bf15.3} & 46.4 \\ \cline{2-10}
        &baseline & 71.5&65.9 & 56.5&47.3 &39.1 &26.0 & 14.3&45.8 \\ 
        &+DDG-Net & {\bf72.5}$^{\textcolor{red}{+1.0}}$&{\bf67.7}$^{\textcolor{red}{+1.8}}$ & {\bf58.2}$^{\textcolor{red}{+1.7}}$&{\bf49.0}$^{\textcolor{red}{+1.7}}$ &{\bf41.4}$^{\textcolor{red}{+2.3}}$ &{\bf27.6}$^{\textcolor{red}{+1.6}}$ & 14.8$^{\textcolor{red}{+0.5}}$& {\bf47.3}$^{\textcolor{red}{+1.5}}$ \\ \hline
\end{tabular}
}
\end{center}
\caption{Comparison results with existing methods on THUMOS14 dataset.}\label{tab:1}
\end{table*}

\begin{table}[htb]
\begin{center}
\resizebox{\linewidth}{!}{
\begin{tabular}{|l|cccc|}
\hline
\multicolumn{1}{|c|}{\multirow{2}{*}{Method}}  & \multicolumn{4}{c|}{mAP(\%)@IoU} \\
        & 0.5 & 0.75 & 0.95 &Avg\\ \hline
        DGAM~\cite{BaifengShi2020WeaklySupervisedAL} &41.0&23.5&5.3&24.4 \\
        RefineLoc~\cite{AlejandroPardo2021RefineLocIR} &38.7&22.6&5.5&23.2 \\
        ACSNet~\cite{ZiyiLiu2021ACSNetAS} &40.1&26.1&{\bf 6.8}&26.0 \\
        HAM-Net~\cite{AshrafulIslam2021AHA}&41.0&24.8&5.3&25.1 \\
        Lee \etal~\cite{PilhyeonLee2020WeaklysupervisedTA}&41.2&25.6&6.0&25.9 \\
        ASL~\cite{JunweiMa2021WeaklySA}&40.2&-&-&25.8 \\
        CoLA~\cite{CanZhang2021CoLAWT}&42.7&25.7&5.8&26.1 \\
        AUMN~\cite{WangLuo2021ActionUM} &42.0&25.0&5.6&25.5 \\
        UGCT~\cite{WenfeiYang2021UncertaintyGC}&41.8&25.3&5.9&25.8 \\
        D2-Net~\cite{SanathNarayan2020D2NetWA} &42.3&25.5&5.8&26.0 \\
        ACM-Net\cite{SanqingQu2021ACMNetAC} &43.0&25.8&6.4&26.5 \\
        CO2-Net~\cite{FaTingHong2021CrossmodalCN}&43.3&26.3&5.2&26.4 \\
        CSCL~\cite{YuanJi2021WeaklySupervisedTA} &43.8&{\bf 26.9}&5.6&26.9\\
        ACGNet~\cite{ZichenYangACGNetAC}&41.8&26.0&5.9&26.1 \\
        Li \etal~\cite{ZiqiangLi2022ForcingTW}&41.6&24.8&5.4&25.2\\
        DGCNN~\cite{HaichaoShiDynamicGM}&42.0&25.8&6.0&26.2\\
        DELU~\cite{MengyuanChenDualEvidentialLF} &44.2&26.7&5.4&26.9 \\\hline
        baseline &43.9&26.6&5.4& 26.8\\
        +DGG-Net&{\bf 44.3}$^{\textcolor{red}{+0.4}}$&{\bf 26.9}$^{\textcolor{red}{+0.3}}$& 5.5$^{\textcolor{red}{+0.1}}$&                   
        {\bf 27.0}$^{\textcolor{red}{+0.2}}$ \\ \hline
\end{tabular}
}
\end{center}
\caption{Comparison results on ActivityNet1.2 dataset.}\label{tab:2}
\end{table}

\section{Experiments}
\subsection{Experimental Setup}
\subsubsection{Datasets}

We evaluate our proposed DDG-Net on two public benchmarks, \ie, THUMOS14~\cite{HaroonIdrees2017TheTC} and ActivityNet1.2~\cite{FabianCabaHeilbron2015ActivityNetAL}.

{
\setlength{\parindent}{0cm}
\textbf{THUMOS14}~\cite{HaroonIdrees2017TheTC} is a challenging dataset that contains 200 validation videos for training and 213 test videos for evaluation from 20 action classes. The videos' length varies from a few seconds to several minutes and each video contains about 15 action instances on average. 
}

{
\setlength{\parindent}{0cm}
\textbf{ActivityNet1.2}~\cite{FabianCabaHeilbron2015ActivityNetAL} is a large-scale dataset that contains 4819 training videos, 2383 validation videos, and 2480 testing videos from 100 action classes. Following previous works, we evaluate our model on the validation set because the ground-truth annotations of the test set are withheld for the challenge.
}

\subsubsection{Evaluation Metrics}
Following standard evaluation protocol, we report mean Average Precision (mAP) at different Intersection over Union (IoU) thresholds.
\subsubsection{Implement Details}
Following existing methods, we use I3D network~\cite{JoaoCarreira2017QuoVA} pretrained on Kinetics dataset~\cite{AndrewZisserman2017TheKH} to extract both the RGB and optical flow features. The action and similarity thresholds are set to 0.5 and 0.8 respectively in default.  
The number of convolution layers \(L\) is 2. 
The temperature \(\tau\)  is set to 0.5. 
The weights \(\lambda_{1}\), \(\lambda_{2}\) are set to 1, 3.2 for THUMOS14 and 0.5, 1.7 for ActivityNet1.2 dataset. 
All experiments are run on a single NVIDIA RTX A5000.

\subsection{Comparisons with State-of-the-Art Methods}\label{5.2}
In Table \ref{tab:1}, we compare the performance of DGG-Net with existing fully-supervised and weakly-supervised  TAL methods on the THUMOS14 dataset. This table shows that our approach outperforms the baseline at all IoU thresholds. Especially mAP has a significant increase of 2.3\% at the IoU threshold of 0.5, and the average of mAP exceeds the baseline of 1.5\%. DGG-Net performs better than all state-of-the-art methods at most IoU thresholds except 0.7 and achieves a new state-of-the-art result with a 0.9\% improvement in the average of mAP. What is more, compared with some fully-supervised and single-frame methods, our approach has competitive and even better performance.

Table \ref{tab:2} presents the comparison results with existing WTAL methods on the AcitivityNet1.2 dataset. The dataset is more simple with an average of 1.5 action instances per video. As shown, our base model has achieved a competitive performance. With DDG-Net, the performance exceeds the base model completely and existing methods at most Iou thresholds. 
It is worth noting that the improvement on ActivityNet 1.2 is slight compared with THUMOS14. This phenomenon is also observed in related work like ACGNet~\cite{ZichenYangACGNetAC}. We attribute the major reason to the difference in the dataset characteristics. THUMOS14 contains 15 action instances per video on average, while ActivityNet1.2 contains only 1.5 action instances. Therefore, THUMOS14 generally contains more ambiguous snippets. According to the result of pre-classification, the proportion of ambiguous snippets is 18.9\% on THUMOS14 and 14.9\% on ActivityNet1.2. Our method contributes to eliminating the adverse effects of ambiguous information, therefore, improves more on THUMOS14. In addition, the performance of pre-classification for pseudo-action and pseudo-background snippets on ActivityNet1.2 is worse, leading to less effective interaction between snippets, which also results in less improvement.

\subsection{Ablation Study}\label{5.3}
{
\setlength{\parindent}{0cm}
{\bf Effects of components.} Table \ref{tab:3} presents the results of adding components on the baseline. The improvement of the average of mAP is 0.6\% when only applying graph average during graph inference. The performance has a sharp decline of 10.3\% because the GCN prefers to generate features attached to action for classification tasks, which is harmful to localization tasks. 
The assimilation of features causes massive false action snippets, which link separate action instances,  leading to the diminution of the number of predicted instances and localization performance. 
With the constraint of feature consistency loss, the model is forced to generate consistent and discriminative features, which significantly improve the performance of the localization module. 
In a word, the result of the localization module is significantly improved with enhanced features, which are generated by DDG-Net equipped with feature consistency loss. 
It is worth noting that the number of parameters increases from 21.5 M to 29.9 M and the number of floating point operations increases from 6.89 G to 13.39 G when equipping the base model with DDG-Net, which is acceptable due to consistent performance gains.

}

\begin{table}[t]
\fontsize{10}{12}\selectfont
\begin{center}
\resizebox{\linewidth}{!}{
\begin{tabular}{|c|c c c|cc|}
\hline
Exp & AVG  & GCN & \(L_{fc}\) & Number & mAP Avg \\ \hline
1 & & & &38118& 45.8 \\
2 & \checkmark& & & 34961&46.4 \\
3 & \checkmark & \checkmark&& 26755&36.1 \\
Ours & \checkmark & \checkmark&\checkmark & 34794&{\bf47.3}\\ \hline
\end{tabular}
}
\end{center}
\caption{Ablation study of the effectiveness of components on THUMOS14. AVG and GCN denote graph average and graph convolution respectively. Number denotes the number of predicted action instances.}\label{tab:3}
\end{table}

\begin{table}[t]
\fontsize{8}{10}\selectfont
\begin{center}
\resizebox{\linewidth}{!}{
\begin{tabular}{|c|c c|cccc|}
\hline
\multicolumn{1}{|c|}{\multirow{2}{*}{Exp}} & \multicolumn{1}{|c}{\multirow{2}{*}{D}} & \multicolumn{1}{c|}{\multirow{2}{*}{FAD}} & \multicolumn{4}{c|}{mAP(\%)@IoU} \\
        &&& 0.1 & 0.4 & 0.7 &Avg\\ \hline
4 & \checkmark& &70  & 46.5&13.4 & 44.9 \\
5 & & \checkmark&71.3  & 47.4&13.6 & 45.8 \\
Ours & \checkmark & \checkmark&{\bf72.5}&{\bf49.0}& {\bf14.8}&  {\bf47.3} \\ \hline
\end{tabular}
}
\end{center}
\caption{Verification of ideas on THUMOS14. D and FAD denote the disconnection of ambiguity and the fused adjacency matrix.}\label{tab:4}
\end{table}

{
\setlength{\parindent}{0cm}
{\bf Analysis of insight.}
Reviewing our idea, ambiguous snippets only should receive information from discriminative snippets and never confuse others with their ambiguity. We denote this strategy in DDG-Net as the disconnection of ambiguity. Additionally, we find it is also essential for keeping collaboration between RGB and Flow features, which refers to the fused adjacency matrix in equation \ref{fuse}. Table \ref{tab:4} demonstrates our experimental results. In Exp 4, we take the respective adjacency matrix used for graph inference. The performance suffers a serious decline because the consistency of the cross-modal features is destroyed after graph inference. Exp 5 shows almost the same performance as the baseline when allowing the propagation of information from ambiguous snippets. In this case, though the complementary information is spread out to relevant snippets, the Ambiguous information reduces the discriminability of features. Only when both the disconnection of ambiguity and the fused adjacency matrix are applied, does our approach generate consistent and discriminative features that improve the performance of the localization module effectively.  Note that our idea can be utilized by other models for modeling relationships between snippets, such as the attention mechanism~\cite{vaswani2017attention}. 
}

\begin{figure}
\begin{center}
\includegraphics[width=1\linewidth]{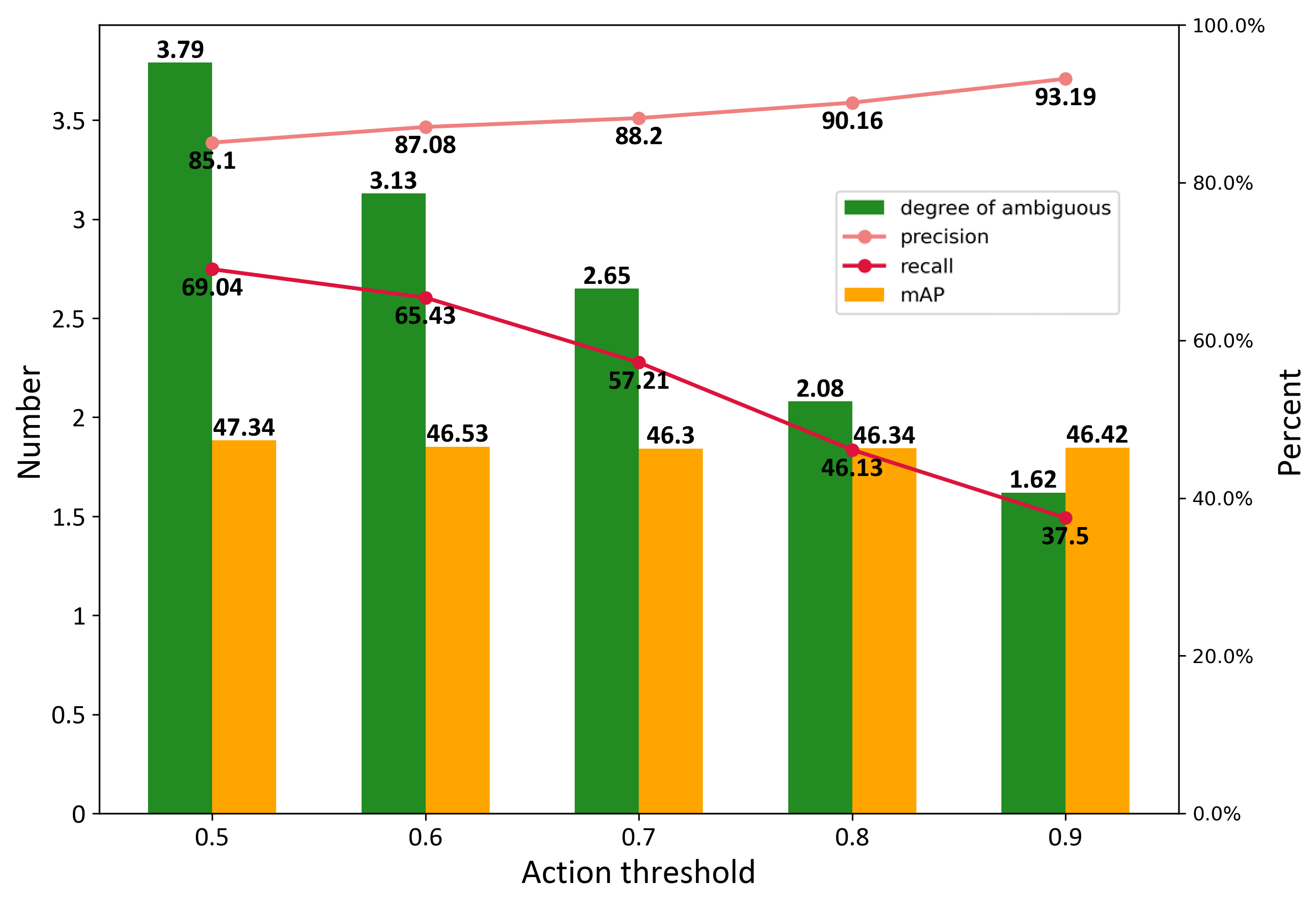}
\end{center}
   \caption{Effects of action threshold.}
\label{fig:3}
\end{figure}

\begin{figure}
\begin{center}
\includegraphics[width=1\linewidth]{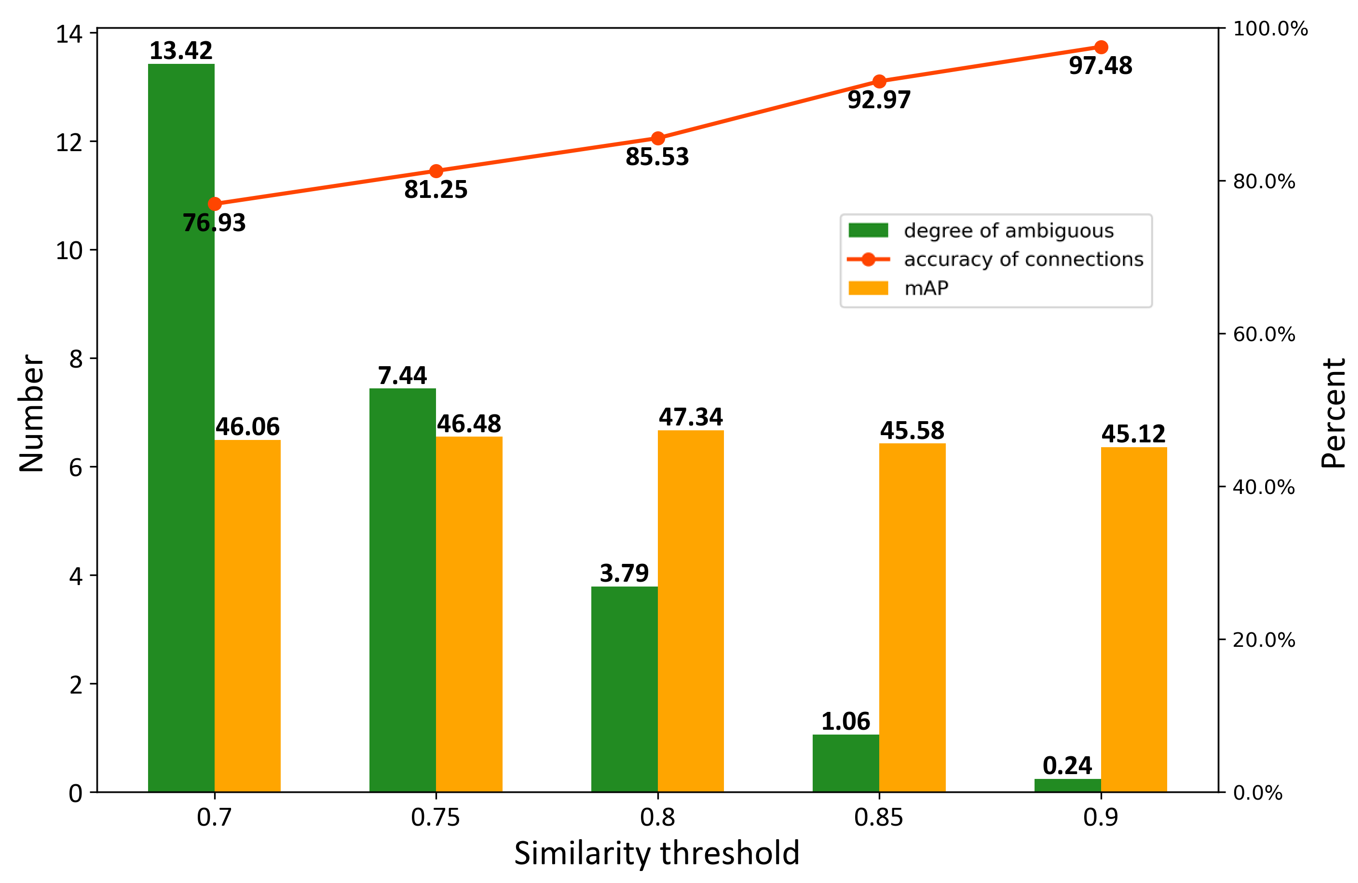}
\end{center}
   \caption{Effects of similarity threshold.}
\label{fig:4}
\end{figure}

{
\setlength{\parindent}{0cm}
{\bf Effects of thresholds.} We first study the effects of the action threshold. Figure \ref{fig:3} shows the result and statistical information at different action thresholds. From the figure, we can find when the action threshold increases, the precision of pre-classification for action and background snippets increases slowly but the recall has a quick decline, which leads to a decrease in the mean degree of ambiguous snippets. Our approach achieves the best balance at the threshold of 0.5. Then we explore the effects of the similarity threshold in Figure \ref{fig:4}. when the similarity threshold increases, the connections for ambiguous snippets are more accurate (the connection is true while linked snippets both belong to action or background) but the mean degree of ambiguous snippets reduces quickly. When the similarity threshold is 0.9, ambiguous snippets receive little discriminative information from few connected snippets, which causes the significant degradation of mAP. Optimal performance is achieved at the threshold of 0.8.

}

\begin{figure*}[htb]
	\centering
	\subfloat[\label{fig5a}]{\includegraphics[width=1\linewidth]{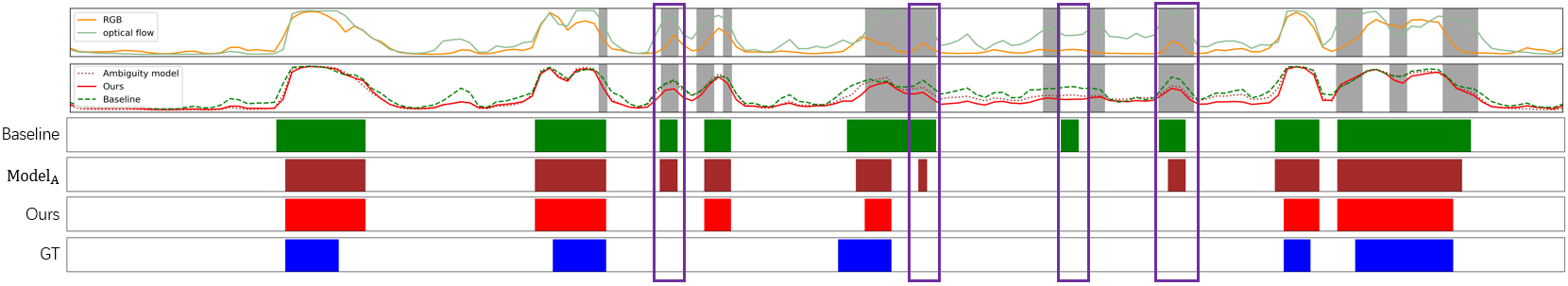}}\hspace{5pt}
	\subfloat[\label{fig5b}]{\includegraphics[width=1\linewidth]{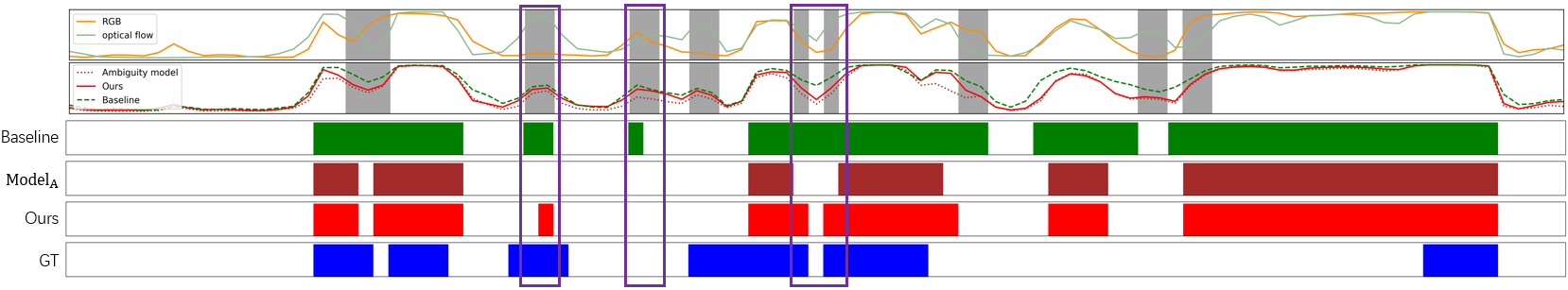}}
	\caption{Visualization results on THUMOS14 dataset. The first and second rows show the original two-stream attention weights and the final attention weights of different models. Following rows present localization results and ground truth (GT). \(\mathrm{Model}_{\mathrm{A}}\) denote the ambiguity model. The \textcolor[RGB]{112,48,160}{purple} boxes show our improvements.}
 \label{fig:5}
\end{figure*}

\begin{figure*}[htb]
\begin{center}
\includegraphics[width=1\linewidth]{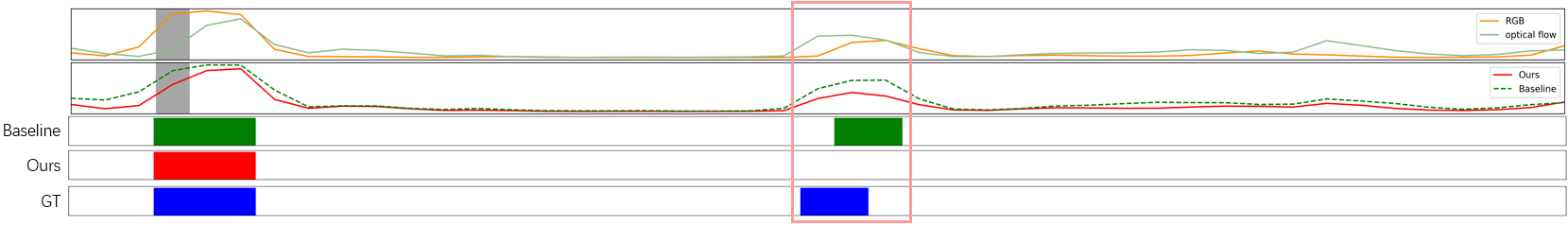}
\end{center}
   \caption{A bad case on THUMOS14 dataset. The \textcolor[RGB]{255,153,153}{pink} box presents a false suppression while corresponding snippets are pre-classify as pseudo-background wrongly.}
\label{fig:6}
\end{figure*}

\subsection{Visualization Results}\label{5.4}
Figure \ref{fig:5} visualizes the qualitative comparisons between the baseline and our method. Additionally, we show the results of Exp 5 in Table \ref{tab:4}, denoted as the ambiguity model. The curves denote the attention weights and shaded parts represent ambiguous snippets. The blocks represent action proposals at the IoU threshold of 0.5. In Figure \ref{fig5a}, we can find both baseline and ambiguity model classify several ambiguous snippets as action snippets which are background actually. Our approach effectively enhances the discriminability of ambiguous snippets and generates more accurate action instances by suppressing the action weights. In Figure \ref{fig5b}, while the ambiguity model suppresses the ambiguous action regions wrongly, DDG-Net has better robustness. 
In addition, we show a bad result compared with the base model in Figure \ref{fig:6}. The action snippets are pre-classified as pseudo-background, which suppresses their attention weights due to receiving background information. It indicates the performance of pre-classification is essential to our method, as mentioned in Section \ref{5.2}.


\section{Conclusions}
In this paper, we present the DDG-Net to explore a novel graph network for effectively enhancing the discriminability of snippet-level representations for WTAL. 
Concretely, we divide snippets using a simple yet valid method and design different types of connections for discriminative and ambiguous snippets. 
In this way, we significantly improve the discriminability of snippet-level representations, especially for ambiguous snippets via graph inference. 
Our method improves the performance of the base model by a large margin and achieves state-of-the-art results. 
Extensive experiments verify the superiority of our method. 

\section*{Acknowledgement}
This work was supported in part by the Major Project for New Generation of AI (No.2018AAA0100400), the National Natural Science Foundation of China (No.62276031, No. 61836014, No. U21B2042, No. 62072457, No. 62006231), the InnoHK program, and the Fundamental Research Funds for the Central Universities.

{\small
\bibliographystyle{ieee_fullname}
\bibliography{paper}
}

\end{document}